\pdfoutput=1
\documentclass[11pt]{article}
\usepackage{acl}
\usepackage{times}
\usepackage{latexsym}
\usepackage[T1]{fontenc}
\usepackage[utf8]{inputenc}
\usepackage{microtype}
\usepackage{graphicx}
\usepackage{cleveref}
\usepackage{subcaption}
\usepackage{hyperref}
\title{Speech Aware Dialog System Technology Challenge (DSTC11)}

\author{Hagen Soltau, Izhak Shafran, Mingqiu Wang, Abhinav Rastogi, Jeffrey Zhao, Ye Jia\\ {\bf Wei Han, Yuan Cao, Aramys Miranda}\\
Google Research\\
\texttt{soltau,izhak,mingqiuwang,arastogi,jeffreyzhao@google.com}\\
\texttt{weihan,yuancao,aramys@google.com, jiaye@tomato.ai}}

\begin{document}
\maketitle
\begin{abstract}
Most research on task oriented dialog modeling is based on written text input. However, users interact with practical dialog systems often using speech as input. Typically, systems convert speech into text using an Automatic Speech Recognition (ASR) system, introducing errors. Furthermore, these systems do not address the differences in written and spoken language. The research on this topic is stymied by the lack of a public corpus. Motivated by these considerations, our goal in hosting the speech-aware dialog state tracking challenge was to create a public corpus or task which can be used to investigate the performance gap between the written and spoken forms of input, develop models that could alleviate this gap, and establish whether Text-to-Speech-based (TTS) systems is a reasonable surrogate to the more-labor intensive human data collection. We created three spoken versions of the popular written-domain MultiWoz task -- (a) TTS-Verbatim: written user inputs were converted into speech waveforms using a TTS system, (b) Human-Verbatim: humans spoke the user inputs verbatim, and (c) Human-paraphrased: humans paraphrased the user inputs. Additionally, we provided different forms of ASR output to encourage wider participation from teams that may not have access to state-of-the-art ASR systems. These included ASR transcripts, word time stamps, and latent representations of the audio (audio encoder outputs). In this paper, we describe the corpus, report results from participating teams, provide preliminary analyses of their results, and summarize the current state-of-the-art in this domain.
\end{abstract}

\section{Introduction}
In recent years, Automatic Speech Recognition (ASR) and Natural Language Processing (NLP) models have converged to utilize common components like Transformers and encoder/decoder modules. They increasingly rely on large amounts of data, large model sizes and large amounts of compute resources. This is a substantial departure from a previous era when ASR and NLP utilized different modeling architectures, chosen to inject domain-specific knowledge and constraints. This shift to a common paradigm has stimulated research in fusing audio and text modalities to substantially improve performance in tasks such as audio intent classification and speech-to-speech translation. The fusion of modalities in many cases allows the direct optimization of the end-task, overcoming the hurdles of the older cascaded approaches that often led to accumulation of errors.

Despite the general trend to develop end-to-end models for various tasks, spoken dialog systems stick out as a sore thumb. Most practical systems utilize a cascaded approach where the output of a general ASR system is fed into a dialog model trained separately on written domain. This mismatch between written and spoken inputs to the dialog models is not well-studied, largely due to the lack of a public task with spoken user inputs.

Research into combined audio-text models is limited by the lack of paired data. While the paired data requirement can be relaxed to some extent for training data via un- or self-supervised training techniques, test sets with paired data are crucial for model evaluation. In addition to an evaluation task, a training set with spoken input would also be helpful in quantifying the gains from supervised learning and recent advances in self-supervised learning.

The focus of our effort was to bring most benefit to the community with the limited resources available. While a Wizard-of-Oz style data collection in spoken domain would have been ideal to fully investigate all the phenomena of spoken domain, that would be extremely labor-intensive especially in annotating the dialog states and was beyond the scope of our effort. Instead, we chose to create a spoken version of a well-studied written-domain task, the MultiWoz task. One advantage of this approach was that we could directly compare the performance of the spoken version with the continuing advances in the written domain. Knowing that the current Text-to-Speech (TTS) systems have come a long way, we also chose to create a TTS-version of the MultiWoz training corpus as a surrogate for the human spoken version.

In the Speech-Aware DSTC11 challenge, participants are asked to infer the dialog states from the sequence of agent (text-input) and user (audio-input) turns. We evaluated the performance on three versions of audio inputs -- TTS-Verbatim, Human-Verbatim (humans speaking the written user inputs), and Human-Paraphrased (humans paraphrasing the written user inputs). Aside from the audio-inputs, we provided transcripts from a state-of-the-art ASR system trained on 33k hours of People's Speech corpus to encourage participation from teams that did not have an easy access to ASR systems. The transcripts were accompanied with timestamps of the words and the latent representations of the acoustic encoder, which could be used by participants to train joint audio-text encoders. 

In the course of developing this challenge, we developed a cascaded baseline system with data augmentation and report performances on a few variants of cascaded systems. In the process, we uncovered a bias in the MultiWoz evaluation task, the slot values in the evaluation set have a substantial overlap with those of the training set. To address this bias, we created a new version of the MultiWoz evaluation set, the dev-dsct11 and eval-dstc11. We observed that the new task surfaces many of the challenges in practical spoken dialog systems associated with mismatch in modalities, inability to recover from ASR errors, and more generally difficulty of extracting semantically relevant information from audio signals.

\section{Related Work}
\label{sec:related}
Motivated by similar consideration as ours, \citet{Kim2022} organized a DST Challenge in 2021 where they created a task using spoken human-human dialogs on tourist information in San Francisco for three
target domains: hotel, restaurant, and attraction. One of the serious limitations of the challenge was that the audio data was not released, only the ASR transcripts. The transcripts had an error rate of about 26.25\% which is significantly higher than the average performance of most state-of-the-art ASR systems. The larger focus of their effort was on evaluating correctness of detecting knowledge-seeking turns, identifying the knowledge snippets and knowledge supplied in the generated responses. Prior to this effort, there have been much smaller efforts with fewer domains and dialogs in DSTC2 and DSTC3~\cite{henderson-etal-2014-second, Henderson2014TheTD}. Here again the organizers only provided ASR transcripts with error rates in high 20s and low 30s, which limited the utility of the corpus as the ASR systems improved over time.

Meanwhile considerable progress has been achieved in improving the naturalness of dialog systems, for example, with chat-bots like Meena~\cite{Adiwardana-meena-2020}, raising expectations of interacting with dialog systems using spoken language. Similarly, the convergence of model architectures for ASR and NLP has stimulated research in creating joint audio-text encoders that could potentially compensate for ASR errors and other spoken language phenomena~\cite{DrexlerG19,Jiawei_NEURIPS2021,chung-2020,tang-etal-2022-unified,slam-2021,mslam-2022,maestro-2022}, where they propose different approaches to align the speech input (frames, phones, utterances) to the corresponding text units. However, none of them have been evaluated on dialog models due to the lack of a dialog task and corpus with audio input. We hope the task and corpus released in this work will bridge this gap along with other recent speech understanding tasks such as superb and slurp~\cite{superb-2021,slurp-2020}


\section{Data}
\label{sec:data}
We chose to create a spoken version of MultiWoz (2.1 version) so we could directly compare the written and spoken versions and reuse the annotations of dialog state labels, avoiding the labor-intensive process of annotating reference labels. 

\subsection{Redesigned DSTC11 Evaluation Sets} \label{sec:dstc11}
Before launching into the data collection and in the process of developing baseline systems, we noticed that there is a substantial overlap in slot values between the training and evaluation sets, leading to overestimation of performance of the models that memorize the slot values, as reported elsewhere~\cite{DSTunseen}. To illustrate the issue, we probed an existing DST model with input whose slot values were replaced with other viable slot values (e.g., {\em cambridge} to {\em new york}, {\em ely} to {\em dublin}), {\em 17:43} to {\em 17:41} and {\em 19:00} to {\em 19:12}). We chose a model that achieves close to the state-of-the-art performance on the task with JGA of 55.4\%~\cite{zhao-etal-2021-effective-sequence}. The model ignores the new slot values and regurgitates the original slot values memorized from the training data, as shown in the Table~\ref{tab:badexamples}.

\begin{table}[h] \centering
\begin{tabular}{|l|l|} \hline
Text Input & Text Output \\ \hline
i want to go to new york & dest=cambridge \\ \hline
i want to go to dublin & dest=ely \\ \hline
train leaving at 17:41 & time=17:43 \\ \hline
train leaving at 19:12 & time=19:00 \\ \hline
\end{tabular}
\caption{Illustration of a model memorizing slot values: the model regurgitates memorized values and ignores the replaced slot values in the input.}
\label{tab:badexamples}
\end{table}

This overlap in slot values between the training and evaluation sets of MultiWoz will mask the effect of  misrecognition of the slot values by ASR systems in a practical spoken dialog system. For a fair evaluation of the models, we redesigned the MultiWoz evaluation sets to replace the slot values in the evaluation sets with new slot values as described below. The replacements were performed at the dialog level to maintain consistency across turns.
\begin{enumerate}
    \item {\em Location Names}: The destination and departure cities for trains and buses were replaced with randomly sampled city names from 12655 cities in the United States.
    \item {\em Hotel Names}: Hotel names were replaced with random names sampled from 1562 hotels in the United States.
    \item {\em Restaurant Names}: Restaurant names were replaced by sampling from 214 restaurants in New York City.
    \item {\em Time Slots}: Timestamps were offset by a random value across all the times mentioned in the dialog.
\end{enumerate}

We measured the impact of the redesigned evaluation sets, henceforth referred to as dev-dstc11 and test-dstc11, using two dialog models -- a seq-to-seq model~\cite{zhao-etal-2021-effective-sequence} and a model that utilizes the power of large language model by fine-tuning the prefix encoding with dialog-specific instructions~\cite{d3st}. The second model has three variants -- D3ST-base, D3ST-large and D3ST-XXL -- related to the size of the underlying T5x large language model. The results, shown in the Table~\ref{tab:unseendst}, confirms our overestimation of the original MultiWoz evaluation sets. The actual performance is almost 50\% worse than reported on the original evaluation sets for most models with the exception of D3ST-XXL. 

\begin{table}[h] \centering
\begin{tabular}{|l|c|c|} \hline
Model & org. Dev & Dev-DSTC11 \\ \hline
seq2seq      &  55.4 & 20.8 \\ \hline
D3ST-base    &  54.2 & 22.0 \\ \hline
D3ST-large   &  54.5 & 25.2 \\ \hline
D3ST-xxl     &  57.8 & 43.1 \\ \hline
\end{tabular}
\caption{Performance of dialog models, measured in JGA, on dev-dstc11, the redesigned MultiWoz evaluation sets. The comparison with original dev set illustrates the substantial overestimation due to the overlap in training and dev slot values.}
\label{tab:unseendst}
\end{table}

The memorization of the slot values completely distorts the effect of ASR errors on the downstream dialog models and paints an overly rosy picture. We illustrate this with a simple cascade baseline where an ASR system transcribes the input speech into written form which is fed into a seq-to-seq dialog model~\cite{zhao-etal-2021-effective-sequence}. In the Table~\ref{tab:seenasrdst} we compare the performance degradation from switching from written to spoken input and then retraining the seq-to-seq model on the ASR transcripts. According to the performance on the original dev set, the degradation from written to ASR transcripts is 58.1\% to 47.2\%. Furthermore, most of the degradation is recovered by retraining on the ASR transcripts. Both these results are incorrect and misleading as will show in the Section~\ref{sec:cascaded}, training on ASR output is not nearly as effective as the results in the table suggests.
\begin{table}[h] \centering
\begin{tabular}{|l|c|c|} \hline
Train Data   &  Test Data  & JGA \\ \hline
Written      &  Written    & 58.1 \\ \hline
Written      &  Spoken/ASR & 47.2 \\ \hline
Spoken/ASR   &  Spoken/ASR & 56.0 \\ \hline
\end{tabular}
\caption{Results of training and evaluating a seq-to-seq model on written and ASR transcripts. The results are misleading as we will show in the Section~\ref{sec:cascaded}.}
\label{tab:seenasrdst}
\end{table}

\subsection{Text-to-Speech Version}
\label{sec:tts-version}
One of the questions we were interested in understanding was whether TTS is a good substitute for speech collected from humans. For answering this question, we generated {\em TTS-Verbatim}, a TTS version of the evaluation test and training sets using the system described in~\cite{Ye2021tts}, a system that represents phonemes and graphemes to represent input text. Additionally, for training data, we generated four versions of the training data, each corresponding to a different TTS speaker. There was no overlap between the speakers in the training and evaluation sets. 

\subsection{Human Data Collection}
\label{sec:human-version}

We focused our data collection on the user turns since the agent turns are already available to any practical dialog system. The data collection was performed via Amazon Mechanical Turk and consisted of two versions -- {\em Human-Verbatim} and {\em Human-Paraphrased}. Crow-workers were presented a full dialog in text including both user and agent turns. In the verbatim version, the workers were instructed to utter the user turns verbatim as naturally as possible, one at a time till the end of the dialog. In the paraphrased version, the workers were instructed to utter a paraphrased version of the user turn preserving the semantic meaning and the entities (e.g., names, times). After they finished the dialog, they were asked to transcribe their own paraphrased user turns. 

The data was collected in several iterations. After each batch was collected, the quality of the recordings were measured and only those above certain quality were retained. There were several factors that contributed to failures: 1) missing audio, 2) incomplete dialog, 3) unintelligible speech, 4) high background noise,  5) speakers uttering verbatim when they should have paraphrased, and 6) speakers not transcribing their utterance in the paraphrased version. 

We developed objective measures to perform quality control in bulk. The missing audio and incomplete dialogs were detected programmatically. To filter out utterances with unintelligible speech and high background noise, we transcribed the collected user utterances with an ASR system (for details of the system see Section~\ref{sec:baseline}) and measured the accuracy with respect to the corresponding transcripts. Since insertions may occur due to disfluencies, we used deletions as a stronger quality indicator and used a threshold of 25\%. 

Measuring the quality of paraphrased version was challenging. We iterated with objective metrics and listen to sampled utterances. After a few iterations, we converged on the following criteria: 1) paraphrased versions had at least 60\% as many words as verbatim, 2) the paraphrased version differed from the verbatim version by at least 45\% as indicated by the WER between recognized words and the written user prompt, and 3) less than 30\% WER with respect to the crowd-worker generated transcripts. Minimum word ratio is to ensure the paraphrased speech carries enough information as in the original written user prompt. The second criterion was to ensure that the paraphrased version was sufficiently different from the written version. The last criterion was to ensure audio quality and intelligibility. 

Besides the automatic quality control, we also sampled and listened to a small portion of the collected recordings. We noticed that even though the crowd-workers were asked to speak naturally, the verbatim was stilted more towards a read version. In a natural dialog, the speakers are likely to show more emotional variations and disfluencies. In comparison, the paraphrased version was more natural since they were using their own words.

In most natural conversations, speakers use 12 hour format for time (e.g., 1315 hrs vs 1:15pm). Since we wanted the crowd-workers to speak as naturally as possible, we replaced all the 24 hour time format in the original MultiWoz set with their corresponding 12 hour format.

\section{Challenge Task and Evaluation}
\label{sec:task}

The main task in the DSTC11 Challenge is to infer the dialog states correctly from the audio inputs that were provided corresponding to the redesigned MultiWoz test set, redesigned as described in Section~\ref{sec:dstc11}. For ablation studies, the participants were required to submit their results for three conditions: {\em TTS-Verbatim}, {\em Human-Verbatim} and {\em Human-paraphrased}, which were collected as described in Section~\ref{sec:tts-version} and~\ref{sec:human-version}. The participants were free to use their own ASR system or the outputs provided from our baseline ASR system, described in Section~\ref{sec:asr}.

The performance in the challenge is measured using the standard Joint Goal Accuracy (JGA) as a primary metric and Slot Error Rate (SER) as a secondary metric for fine-grained comparisons. In the literature, results are often difficult to compare since research groups apply their own output normalization before scoring the results. To sidestep the resulting confusion, this challenge is evaluated using standard evaluation tool published on GitHub~\cite{zang-etal-2020-multiwoz}. We created a patch to support the results from this challenge and published the golden reference for our redesigned evaluation sets, which can be downloaded from the DSTC11 challenge website~\url{https://dstc11.dstc.community}.

The participants were allowed to utilize any publicly available data or checkpoints, but no private data to allow fair comparisons. They were allowed to augment data and no restrictions were imposed on model sizes or computational costs.
\section{Baseline}
\label{sec:baseline}

\subsection{ASR and Related Outputs} \label{sec:asr}
As mentioned earlier, we provided the output of a baseline ASR system to reduce the barrier for participation by teams which didn't have ASR systems available to them.

\begin{figure}[h]
    \centering
    \includegraphics[width=6cm]{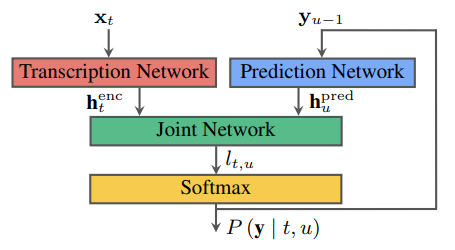}
    \caption{RNN-Transducer}
    \label{fig:rnnt}
\end{figure}

The baseline ASR model is based an RNN-Transducer~\cite{Graves2012} which is essentially an encoder-decoder model, as illustrated in Figure~\ref{fig:rnnt}. Unlike typical encoder-decoder models, the RNN-T uses hard attention and learns alignments via Forward-Backward algorithm similar to HMM, albeit with a different topology. These models have been extremely successful in ASR and achieves state-of-the-art performance in several benchmarks.

Our baseline models is trained on the PeopleSpeech corpus~\cite{PSdata}, a supervised, publicly available data set of approx. 32,000 hours of audio data. The encoder consists of 16 Transformer layers, and the language model (decoder) is a single LSTM layer, altogether a 220m parameter model. The performance of the model on the MultiWoz evaluation sets are reported in Table~\ref{tab:asr-wer}.
\begin{table}[h] \centering
   \begin{tabular}{|l|c|c|} \hline
                &   TTS-Verbatim      &   Human-Verbatim  \\ \hline
    dev-dstc11  &   8.1(5.7/0.3/2.1)  & 11.9(7.7/2.7/1.5) \\ \hline
    test-dstc11 &   8.2(5.7/0.3/2.1)  & 13.0(8.2/3.4/1.5) \\ \hline
   \end{tabular}
   \caption{Performance of the baseline ASR model on the MultiWoz evaluation sets, reported as WER(Sub/Del/Ins)} 
   \label{tab:asr-wer}
\end{table}

We provided the following outputs as part of the DSTC11 challenge corpus.
\begin{enumerate}
    \item {\em Audio Waveforms (16Khz/16-bit PCM)}: This allows researchers to investigate DST models that directly operate on audio such as end-to-end systems.
    \item {\em Audio Encodings (75ms frame rate, 1024-dim)}: These are the output activation from the last Transformer layer in the audio encoder. Since the input to the model are log-mel features at 25ms frame rate, and the encoder reduces the frame rate by 3x, the resulting output frame rate is 75ms, i.e. for an utterance of 15 sec, the audio encodings would be of length 200. These encodings are provided to support research in loosely-coupled ASR-DST cascaded systems. 
    \item {\em ASR Hypotheses with Time Alignment}: The ASR recognition outputs are provided to support research in different cascaded ASR-DST systems. Time alignments allow teams to pinpoint the location of recognized tokens in the audio encodings. Note, due the nature of RNN-T, multiple output tokens (words) can correspond to the same audio frame.
\end{enumerate}

Typical recognition errors (ordered by frequency), illustrated in Table~\ref{tab:asr-errors}, include time formatting issues, spoken single-digit numbers, split words, and more general misspelled content words, where the last category is probably very relevant for DST purposes.
\begin{table}[h] \centering
   \begin{tabular}{|l|l|} \hline
   \multicolumn{2}{|c|}{Spoken vs Written Numbers} \\ \hline
    3 stars & three stars \\ 
    031 & thirty-on \\ \hline
    \multicolumn{2}{|c|}{Split Words} \\ \hline
    i am & i'm \\
    wifi & wi-fi \\
    guesthouse & guest house \\
    goodbye & good bye \\
    double-checking & double checking \\ \hline
     \multicolumn{2}{|c|}{Misspelled content words} \\ \hline
    hennessey &  hennessy \\
    moxee & marcy \\
    kinbrae & kinbray \\ \hline     
   \end{tabular}
   \caption{Typical ASR Recognition Errors} 
   \label{tab:asr-errors}
\end{table}

\subsection{Dialog State Tracking Models}
\label{sec:dst}
For our baseline DST experiments, we relied on two models.
\begin{enumerate}
    \item Vanilla seq-to-seq model: This seq-to-seq model is obtained by fine-tuning a T5-base model on MultiWoz corpus~\cite{zhao-etal-2021-effective-sequence}.
    \item D3ST: This is also a seq-to-seq model trained on T5, but the input to this model contains a prompt which describes slot names in short natural language descriptions along with potential values~\cite{d3st}. The ordering of the slot names are randomizes to improve robustness.
\end{enumerate}

\subsection{Data Augmentation}
Data augmentation is a common technique to improve accuracy and robustness. Since our preliminary results in Table~\ref{tab:unseendst} unearthed problems related to memorization, we felt the need to incorporate data augmentation into our baseline systems. We created new versions of user responses for training data by replacing slot values with randomly picked city names, time offsets and restaurant names, as described in Section~\ref{sec:dstc11}. The new slot names were drawn from a different list than the ones used for generating the redesigned evaluation sets. In all, we generated about 100x training data.

\subsection{Cascaded ASR-DST System} \label{sec:cascaded}
For a baseline, we created a cascades ASR-DST system where the transcripts from the ASR model in Section~\ref{sec:asr} were fed as input to the DST model in Section~\ref{sec:dst}. The performance of the models of various sizes are summarized in Tables~\cref{tab:dstbase,tab:d3stbase,tab:d3stlarge,tab:d3stxxl}. The best baseline performance is obtained with D3ST-xxl, obtaining 33.5\% JGA on the human paraphrased test set. 
\begin{table} \centering
   \begin{tabular}{|l|c|c|c|c|} \hline
   Test Data & \multicolumn{4}{|c|}{Training Data} \\ \hline
   & \multicolumn{2}{|c|}{Text (Ref)} &  \multicolumn{2}{|c|}{ASR Hyp} \\ \hline
   & 1x & 100x & 1x & 100x \\ \hline
   Text        & 20.1 & 41.7 & - & - \\ \hline
   H-Paraphrased & 16.5 & 24.1 & 18.3 & 25.6 \\ \hline
   \end{tabular}
   \caption{ASR-Vanilla Seq-to-Seq DST {\bf base} cascaded model (JGA)} 
   \label{tab:dstbase}
\end{table}

\begin{table}[h] \centering
   \begin{tabular}{|l|c|c|c|c|} \hline
   Test Data & \multicolumn{4}{|c|}{Training Data} \\ \hline
   & \multicolumn{2}{|c|}{Text (Ref)} &  \multicolumn{2}{|c|}{ASR Hyp} \\ \hline
   & 1x & 100x & 1x & 100x \\ \hline
   Text     & 22.0 & 41.6 & 20.9 & 28.6 \\ \hline
   TTS-Verbatim & 19.2 & 27.7 & 19.9 & 27.1 \\ \hline
   H-Verbatim & 17.1 & 23.6 & 17.6 & 22.5 \\ \hline
   H-Paraphrased & 17.0 & 22.8 & 17.3 & 21.8 \\ \hline
   \end{tabular}
   \caption{ASR-D3ST {\bf base} cascaded model (JGA)} 
   \label{tab:d3stbase}
\end{table}

\begin{table}[h] \centering
   \begin{tabular}{|l|c|c|c|c|} \hline
   Test Data & \multicolumn{4}{|c|}{Training Data} \\ \hline
   & \multicolumn{2}{|c|}{Text (Ref)} &  \multicolumn{2}{|c|}{ASR Hyp} \\ \hline
                 & 1x & 100x & 1x & 100x \\ \hline
   Text          & 25.2 & 44.2 & 22.3 & 32.6 \\ \hline
   TTS-Verbatim  & 20.5 & 29.1 & 20.7 & 29.5 \\ \hline
   H-Verbatim    & 18.6 & 25.1 & 18.5 & 24.8 \\ \hline
   H-Paraphrased & 18.0 & 23.9 & 17.5 & 23.6 \\ \hline
   \end{tabular}
   \caption{ASR-D3ST {\bf large} cascaded model (JGA)} 
   \label{tab:d3stlarge}
\end{table}

\begin{table}[h] \centering
   \begin{tabular}{|l|c|c|c|c|} \hline
   Test Data & \multicolumn{4}{|c|}{Training Data} \\ \hline
   & \multicolumn{2}{|c|}{Text (Ref)} &  \multicolumn{2}{|c|}{ASR Hyp} \\ \hline
                 & 1x & 100x & 1x & 100x \\ \hline
   Text          & 43.1 & 52.8 & 33.9 & 43.0 \\ \hline
   TTS-Verbatim  & 27.3 & 32.1 & 26.8 & 38.4 \\ \hline
   H-Verbatim    & 23.6 & 27.9 & 23.7 & 31.8 \\ \hline
   H-Paraphrased & 21.8 & 26.1 & 21.9 & 30.9 \\ \hline
   \end{tabular}
   \caption{ASR-D3ST {\bf xxl} cascaded model (JGA)} 
   \label{tab:d3stxxl}
\end{table}

We point a few pertinent observations on cascaded ASR-DST models and data augmentation.
\begin{enumerate}
\item TTS (TTS-Verbatim) shows a degradation with respect to written version. However, the degradation is larger with Human-Verbatim (41.0\% vs. 33.5\% JGA), confirming our suspicion that we cannot rely on TTS as a surrogate for human speech.
\item Surprisingly, the drop in performance from Human-Verbatim to Human-Paraphrased is not very large. This is not because the two versions are similar since quality checks described in Section~\ref{sec:human-version} assures us that the paraphrased version is sufficiently different from verbatim version.
\item For models of smaller sizes, the differences in model type (Vanilla seq-to-seq vs D3ST) seems less relevant. However, scaling the D3ST model from D3ST-Large (3B params) to D3ST-XXL (11B params) has substantial impact on performance (24.8\% vs 33.5\% JGA). This raises a question of fairness since teams may not have resources to work with larger model sizes. Nonetheless, we did find examples in results provided by participants were smaller models were able to outperform larger models, see Section~\ref{sec:results} for details.
\item Training DST models with ASR hypotheses is simple and improves the performance substantially from 28.2\% to 33.5\% even when the training data is based on TTS and not human speech.
\item Data augmentation gives consistent gains across all conditions.
\end{enumerate}

\section{Results}
\label{sec:results}

\begin{figure}[h]
    \centering
    \includegraphics[width=0.4\paperwidth]{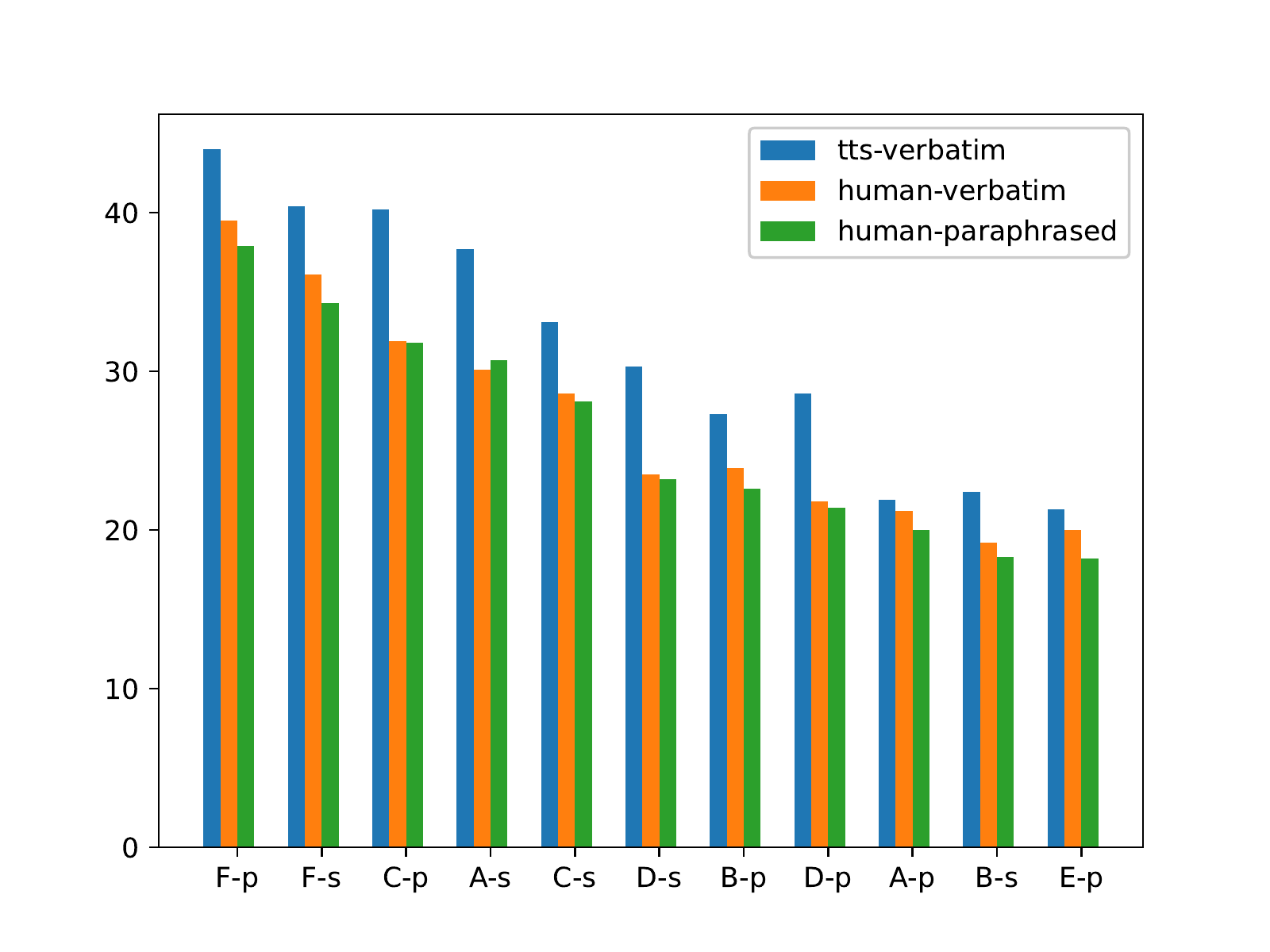}
    \caption{Joint Goal Accuracy (JGA) of team's submissions}
    \label{fig:jga}
\end{figure}

\begin{figure}[h]
    \centering
    \includegraphics[width=0.4\paperwidth]{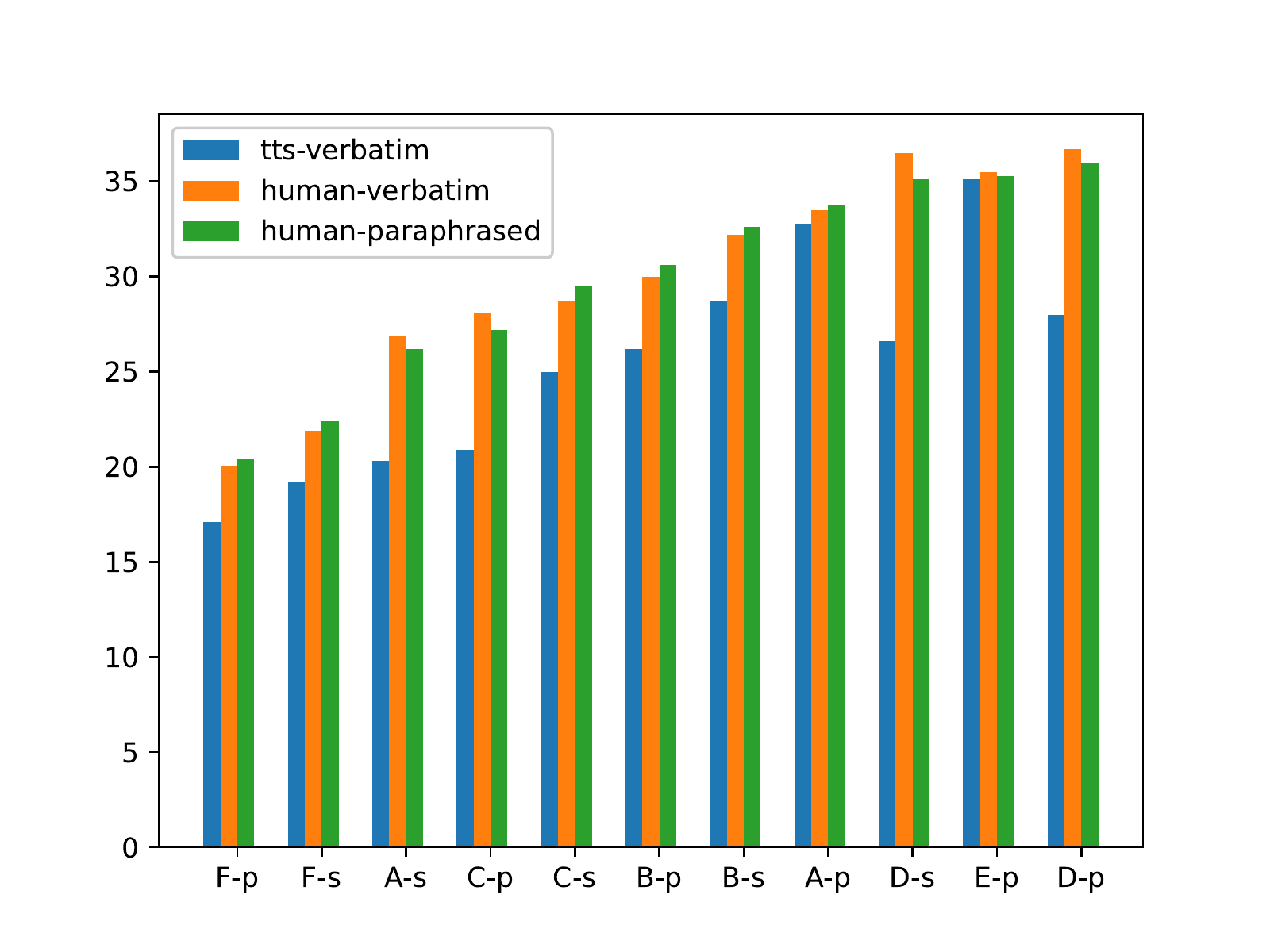}
    \caption{Slot Error Rate (SER) of team's submissions}
    \label{fig:ser}
\end{figure}

We received $11$ submissions from $6$ teams and the results are shown in figures~\ref{fig:jga} and~\ref{fig:ser}. The full set of numbers are also available in the link under the DSTC11 Challenge website \url{https://dstc11.dstc.community}.

We provide a high level overview of the results, but refer to the team's system descriptions for details. While we provided data to build direct audio-to-dst and tightly coupled models, all teams chose a cascaded approach with separately trained ASR and DST models. Many teams employed an explicit ASR error correction model and re-trained their DST models on ASR hypotheses together with various forms of TTS-based data augmentation. Within this general approach teams experimented with several variations and as a result the performance across submissions vary substantially. The highest performing submission obtains a JGA of 37.9\% while the lowest performance is at 18.2\%.

\subsection{Alternative ASR models}
While we provided ASR output with our baseline ASR system, described in Section~\ref{sec:asr} based on {\em PeopleSpeech} corpus, three submissions used {\em Whisper} instead~\cite{radford2022robust}. We compared the two ASR models to tease apart the differences. As reported in Table~\ref{tab:asr-wer-whisper}, not surprisingly, we found that {\em Whisper} transcribes the evaluation sets more accurately than our baseline model (see Table~\ref{tab:asr-wer}) since it is trained on a magnitude order more data.
\begin{table}[h] \centering
   \begin{tabular}{|l|c|c|} \hline
         &   TTS-Verbatim             &   Human-Verbatim \\ \hline
    dev-dstc11  &   4.8(3.8/0.6/0.4)  & 8.5(5.8/1.4/1.4) \\ \hline
    test-dstc11 &   4.6(3.7/0.6/0.3)  & 8.9(6.1/1.5/1.3) \\ \hline
   \end{tabular}
   \caption{Performance of Whisper on evaluation sets, reported as WER(Sub/Del/Ins).} 
   \label{tab:asr-wer-whisper}
\end{table}

We evaluated {\em Whisper} in the cascaded ASR-DST system with two models, one trained on 100x augmented written text and one trained on {\em Whisper} transcripts of the TTS version of the same training data. The results shown in Table~\ref{tab:d3stxxl-whisper} clearly demonstrate that the improvements in transcription accuracy translates to improvement in DST accuracy, whereby JGA scores improve from $30.9\%$ to $34.3\%$ for the Human-Paraphrased test data.

\begin{table}[h] \centering
   \begin{tabular}{|l|c|c|} \hline
   Test Data & \multicolumn{2}{|c|}{100x Training Data} \\ \hline
                 & Text (Ref) &  ASR Hyp \\ \hline
   Text          & 52.7 & 45.3 \\ \hline
   TTS-Verbatim  & 35.6 & 41.3 \\ \hline
   H-Verbatim    & 32.3 & 35.5 \\ \hline
   H-Paraphrased & 30.9 & 34.3 \\ \hline
   \end{tabular}
   \caption{Performance in JGA of cascaded Whisper ASR-D3ST-XXL trained on text and Whisper hypotheses, evaluated on Whisper hypotheses of evaluation sets.} 
   \label{tab:d3stxxl-whisper}
\end{table}

\section{Conclusions}
In this paper, we describe a new corpus for stimulating research in modeling spoken dialogs that builds on popular written dialog corpus, MultiWoz. Having a spoken version of the same evaluation set allows researchers to study and bridge the performance gap between written and spoken dialog models. In the course of characterizing the task, we observed that there is substantial overlap between the slot values in the training and evaluation sets of the original MultiWoz corpus. We redesigned the evaluation sets (dev-dstc11 and eval-dstc11) by sampling new non-overlapping slot values and show that the new sets captures the weakness of the written dialog models better. We released three versions of the task -- TTS-Verbatim, Human-Verbatim and Human-Paraphrased, where the last two were collected from crowd-workers. One caveat of this collection is that disfluencies such as speech repair are not as well represented as in natural conversations.

We have characterized this gap using a baseline cascaded ASR-DST system using ASR and DST models. While the performance improves with model size and data augmentation, even the best models (D3ST-XXL) show substantial drop in performance when switching from written version to the spoken version (53.8\% to 26.1\% JGA). Retraining the D3ST-XXL model on the ASR hypotheses, improves the performance to 30.9\% JGA but still leaves substantial ground to be covered. 

We report the results of 11 submissions from the DSTC11 Challenge. The dominant paradigm across teams was to rely on large language models for DST. Several submissions inserted ASR error correction modules of different complexities. The performance of the teams varied from a high of 37.9\% to a low of 18.2\% JGA. 

One area of research that is not well explored is better utilization of the latent representations of the audio encoder in the ASR in the downstream DST models. Similarly, we hope the release of the audio and the audio encoders outputs will allow researchers to evaluate the power of joint audio-text encoders on dialog tasks.


\bibliography{paper}

\end{document}